\def\year{2022}\relax
\documentclass[letterpaper]{article} 
\usepackage{aaai22}
\usepackage{times}
\usepackage{helvet}
\usepackage{courier}
\usepackage[hyphens]{url} 
\usepackage{graphicx} 
\def\UrlFont{\rm}
\usepackage{natbib}
\usepackage{multibib}
\usepackage{caption} 
\DeclareCaptionStyle{ruled}{labelfont=normalfont,labelsep=colon,strut=off} 
\usepackage{amsmath,graphicx}
\usepackage{amsthm,amsmath,amssymb}
\usepackage{mathrsfs}
\usepackage{booktabs}
\usepackage{multirow}
\usepackage{tabularx}
\usepackage{graphicx}
\usepackage{amsfonts}
\usepackage{threeparttable}
\usepackage{algorithm}
\usepackage{algorithmic}
\usepackage{appendix}
\usepackage{newfloat}
\usepackage{listings}
\floatstyle{ruled}
\newfloat{listing}{tb}{lst}{}
\floatname{listing}{Listing}

\title{MoEC: Mixture of Expert Clusters}
\author{
    Yuan Xie\textsuperscript{\rm },
    Shaohan Huang\textsuperscript{\rm },
    Tianyu Chen\textsuperscript{\rm },
    Furu Wei\textsuperscript{\rm }
}
\affiliations{
    \textsuperscript{\rm}Microsoft Research Asia, China\\
    \{v-yuanxie, shaohanh, v-tianyuchen, fuwei\}@microsoft.com
    
}

\begin{document}

\maketitle

\begin{abstract}
Sparsely Mixture of Experts (MoE) has received great interest due to its promising scaling capability with affordable computational overhead. MoE converts dense layers into sparse experts, and utilizes a gated routing network to make experts conditionally activated. However, as the number of experts grows, MoE with outrageous parameters suffers from overfitting and sparse data allocation. Such problems are especially severe on tasks with limited data, thus hindering the progress for MoE models to improve performance by scaling up.
In this work, we propose \textit{Mixture of Expert Clusters} — a general approach to enable expert layers to learn more diverse and appropriate knowledge by imposing variance-based constraints on the routing stage.
We further propose a cluster-level expert dropout strategy specifically designed for the expert cluster structure. 
Our experiments reveal that MoEC could improve performance on machine translation and natural language understanding tasks, and raise the performance upper bound for scaling up experts under limited data. We also verify that MoEC plays a positive role in mitigating overfitting and sparse data allocation.

\end{abstract}

\section{Introduction}

Scaling up the model capacity has shown to be promising to achieve better performance on a variety of tasks, including natural
language understanding~\cite{brown2020language,raffel2019exploring} and visual representation learning~\cite{dosovitskiy2020image,bao2021beit}. The continued growth in model size and parameters brings higher computational cost, while large dense models have almost hit the boundary of hardware capacity. In pursuit of better computational efficiency, sparse Mixture-of-Experts (MoE) is proposed as an efficient alternative to dense models~\cite{lepikhin2020gshard,fedus2021switch,riquelme2021scaling,lewis2021base}. For the sparsely-gated MoE transformers, the feed-forward network (FFN) sub-layer will be replaced by a set of experts with independent parameters.

The sparsity of MoE is brought by experts and the gated routing network. The gated routing network will calculate the routing score between input tokens and each expert and activate experts with top-k routing scores. Most experts will not be activated, thus forming a sparse structure. Since the computation cost is only proportional to the activated top-k sub-network, sparsely activated MoE models could scale up model parameters without significantly increasing computational cost. With affordable computational overhead, MoE models could achieve better performance than dense models on various tasks such as neural machine translation~\cite{lewis2019bart,conneau2019cross,lepikhin2020gshard},image recognition~\cite{riquelme2021scaling} and speech recognition~\cite{kumatani2021building}.

\begin{figure}
    \centering    \includegraphics[width=\linewidth]{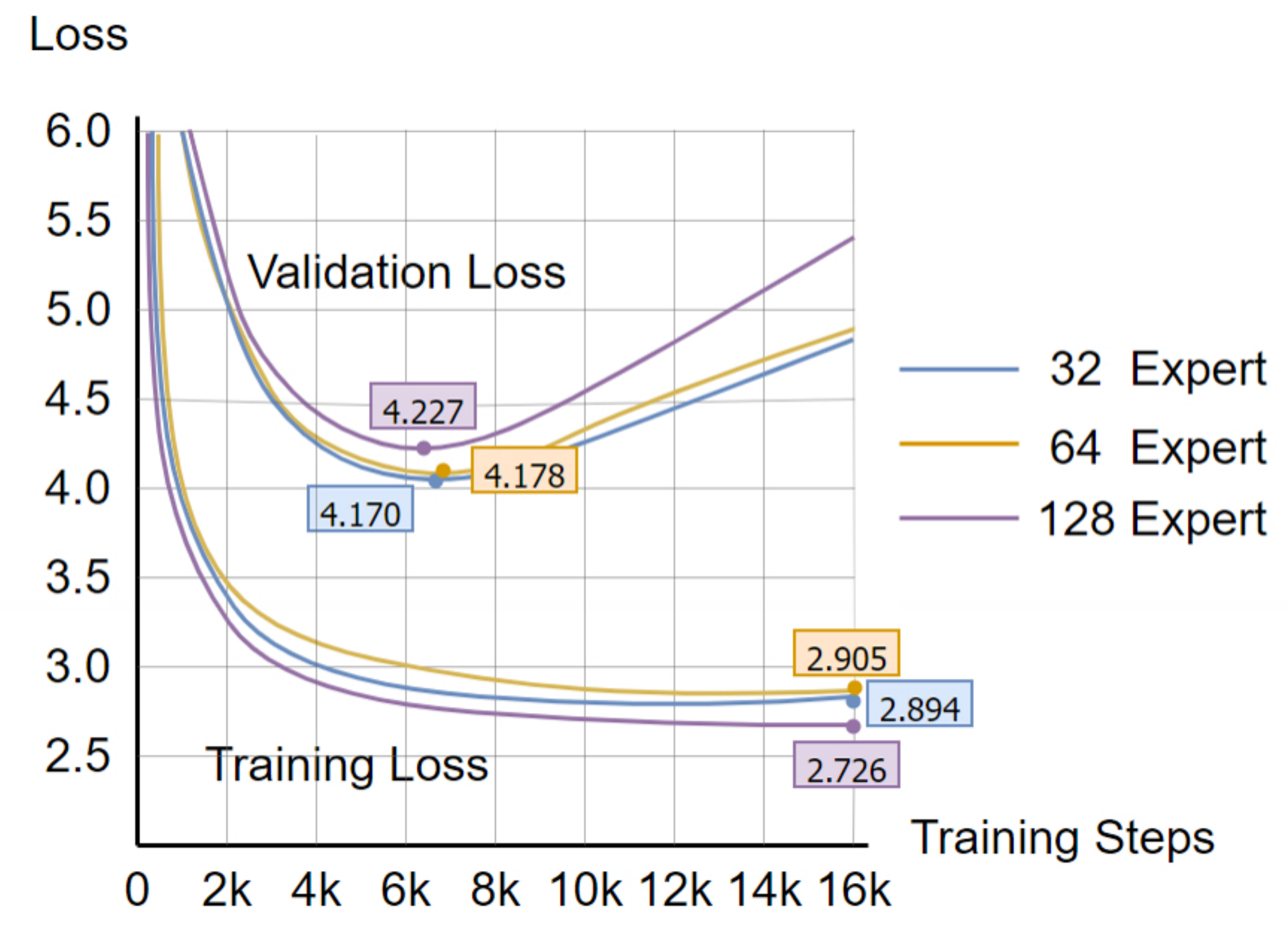}
    \caption{A simple demonstration of loss curves of MoE models on WMT-14 English-to-German translation task. We show the loss curve of MoE baseline models with different experts. The value in the box represents the minimum loss.} 
    \label{fig:overfit}
\end{figure}

Recent studies have reached a consensus that more experts mean more parameters and large model capacity, which always bring improvements. However, some studies show that more trainable parameters and sparse conditional computation may introduce overfitting~\cite{xue2021go,lou2021sparse,xue2022one}, especially for downstream tasks with limited data. As depicted in Figure~\ref{fig:overfit}, as the number of experts grows, overfitting gradually becomes apparent in the machine translation task. Moreover, we find that enlarging the size of the MoE will not always lead to improvement. There seems to exist a performance upper bound of scaling up experts with limited data.

Moreover, we find an unreasonable phenomenon in Figure~\ref{fig:overfit}: 64-expert MoE with more parameters and larger model capacity has higher training loss than 32-expert MoE. It implies that large-scale MoE not only suffers from overfitting, but also has other hidden problems that affect training.
According to our analysis, the probability of each expert getting a token reduces proportionally as the number of experts grows. With the same data, each expert will get less diverse samples. It may affect the sufficient learning of expert layers. Insufficient data to match more parameters is also a major cause of overfitting. Therefore, we want to explore ways in which experts could get diverse samples and learn abundant knowledge, thereby alleviating overfitting and sparse data allocation.

In this work, we propose Mixture of Expert Clusters (MoEC), a general optimizing strategy for MoE models. We close the routing probability among neighbor experts to form the clustered expert structure. The inductive bias expects that the similarity of intra-cluster experts is high while the similarity of inter-cluster experts is low. Experts within a cluster are prone to tokens with similar hidden states and could ``share'' similar tokens. Moreover, we propose a cluster-level expert dropout strategy for the expert cluster structure. Several experts in the cluster will be randomly dropped, the dropped experts will not participate in the routing stage. The activated experts will be selected from the remaining experts in the cluster. Implementing dropout within clusters will make tokens always dispatched to suitable experts, no matter how random the dropout is.

We evaluate our MoEC on machine translation and natural language understanding tasks. Experiment results show that MoEC outperforms dense models and baseline MoE models. It indicates that MoEC retains the advantages of the sparse structure of MoE, and alleviates overfitting and sparse data allocation problems.

Our contributions are summarized as follows:

\begin{itemize}
    \item We point out the overfitting and sparse data allocation problems for large-scale MoE models, and experts getting less diverse samples could be the common cause of both problems.

    \item We propose to build expert clusters by variance-based constraints, which allows experts to get a more diverse set of similar tokens. We also implement cluster-level expert dropout as a regularization method.
    
    \item We conduct experiments on translation and natural language understanding tasks. MoEC could improve performance and alleviate problems caused by scaling up experts.
    
    \item We find that there exists a performance upper bound for scaling up MoE models with limited data, and MoEC could raise the upper bound.
    
\end{itemize}

\section{Related Work}

In the context of modern deep learning architectures, scaling up transformers using sparse Mixture of Experts (MoE) is proven to be effective to achieve state-of-the-art performance on various NLP and CV tasks~\cite{shazeer2017outrageously,lepikhin2020gshard,riquelme2021scaling,fedus2021switch}. Compared with dense transformers, an MoE model contains several experts (feed-forward networks), and a router to select top-k experts for input tokens. It increases the model capacity by such conditional computation while maintaining computational efficiency. To future explore the potential of MoE, some studies focus on router assignment algorithm~\cite{lewis2021base,roller2021hash,dai2022stablemoe}. Besides, some work focus on optimizing training methods for MoE models.~\citet{dua2021tricks} applied a temperature heating mechanism for sparse MoE models on the translation task.~\citet{chi2022representation} proposed a dimension reduction to estimate the routing scores between tokens and experts on a low-dimensional hyper-sphere. Our work is also proposed to optimize the MoE model. Instead of changing the model structure and routing strategy, MoEC establishes expert clusters, which allows experts to be
assigned a more diverse set of similar tokens.

Although MoE models have achieved promising results, they are proven to have overfitting problems~\cite{fedus2021switch,wu2022residual,xue2022one} on downstream tasks with limited data. To mitigate overfitting, some works use knowledge distillation to distill MoE models into small-sized MoE models or dense models~\cite{xue2022one,dai2022stablemoe}. Another approach is to apply the dropout strategy during training.~\citet{fedus2021switch} set a small dropout rate at non-expert layers and a larger dropout rate at expert layers.~\citet{liu2022gating} propose gating dropout, which allows some tokens to ignore the gated routing network and stay at their local machines to reduce cross-machine communication. In our work, we propose the cluster-level expert dropout. Randomly selected experts in the cluster will be dropped so that they will not participate in the routing stage.

\section{Preliminary}
\label{sec:preliminary}

To build MoE transformers, it is a common practice to replace feed-forward network (FFN) sub-layers with a set of experts. The experts share the same structure with the FFN layer in the dense transformer model. We denote the hidden representation of input token $x$ as $h$, and the embedding for the $i$-th expert as $e_i$. The router computes the routing score $s_i=h^\mathrm{T}e_i$ to compare the similarity between $h$ and the set of experts $E$. Then, the router utilizes a gating function $\alpha(\cdot)$ to compute the gated value of expert $i$.

\begin{equation}
    \alpha_i = \left\{
        \begin{aligned}
        \frac{exp(s_i)}{\sum_{j=1}^{E}exp(s_j)},\quad \textit{softmax}\ gating  \\
        \frac{1}{1+exp(-s_i)},\quad \textit{sigmoid}\ gating
    \end{aligned}
    \right.
\end{equation}

The gating function $\alpha_i$ represents the probability of dispatching input token to expert $i$. The top-k gated-value is used for dispatching the token $x$ according to $\alpha_i$. The corresponding k expert networks are conditionally activated. We denote the set of selected top-k indices as $K$.

\begin{equation}
    y=\sum_{i\in K} \alpha_i \cdot E_i(x)
\end{equation}

where $E_i(x)$ is the $i$-th expert network, which is a feed-forward network. The output of the gated routing network is the linearly weighted combination of each expert’s computation on the token by the gate value.

\section{Method}
\label{sec:Method}
In this work, our goal is to give experts access to more diverse training samples, thus learning abundant knowledge and mitigating overfitting and sparse data allocation. We close the routing probability among neighbor experts to form the clustered expert structure. We apply the variance-based clustering loss to implement constraints. Then, we further propose a cluster-level expert dropout strategy. In our work, we use top-1 gating. Only the expert with the largest routing score is activated. And we choose softmax as our activation function. Experts in a cluster will be distributed on the same device to reduce communication costs.

\subsection{Mixture of Expert Clusters}

\begin{figure}
    \centering
    \includegraphics[width=\linewidth]{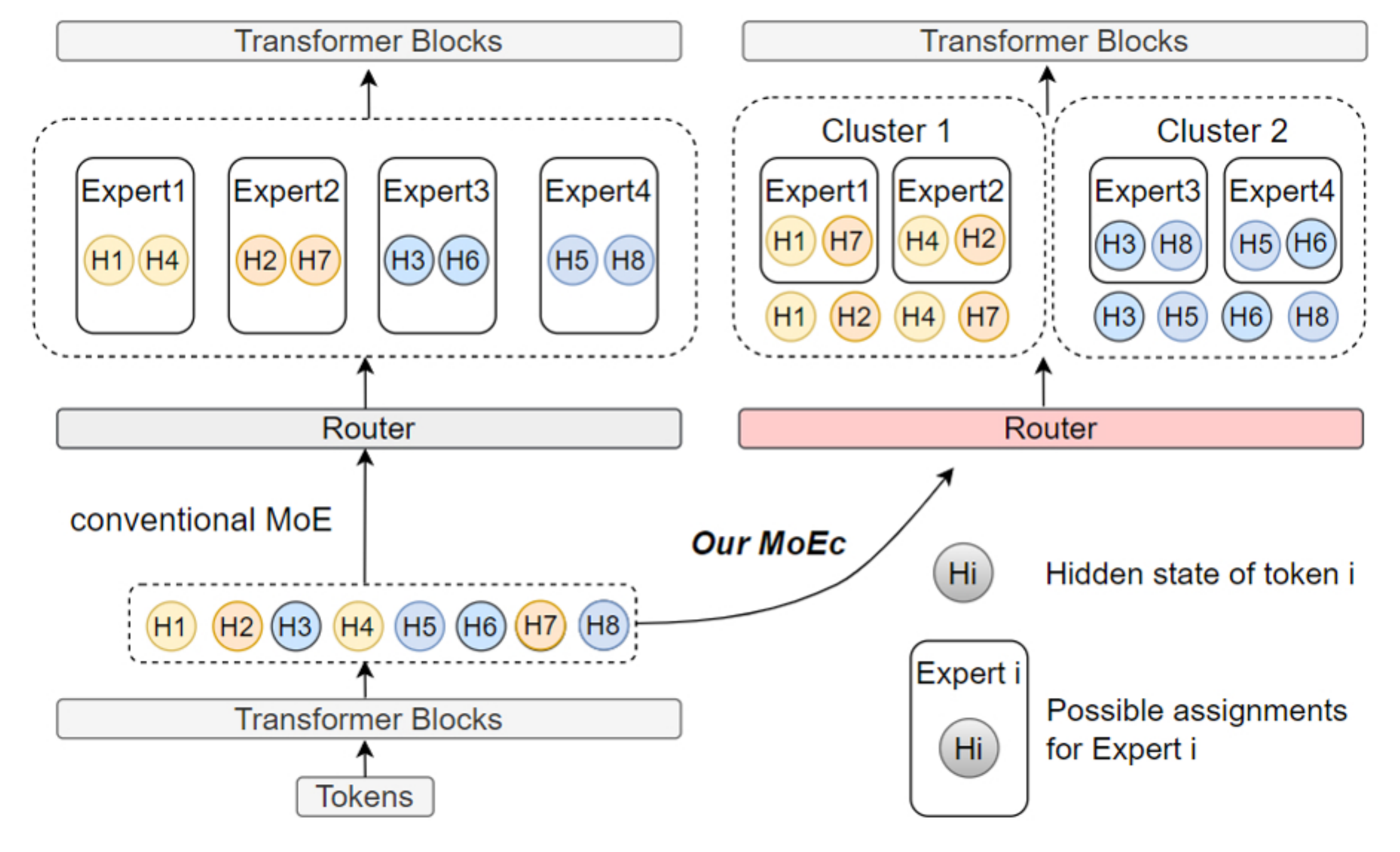}
    \caption{Illustration of a conventional MoE layer and our proposed MoEC layer. The similarity between hidden states $H_i$ is represented by the color.} 
    \label{fig:overview}
\end{figure}

We illustrated our MoEC (Mixture of Expert Clusters) in Figure~\ref{fig:overview}. For conventional MoE, the routing probability of tokens will not be constrained. The router will always dispatch input tokens to their best-matched experts, while other similar tokens have little chance of being selected. When scaling up the number of experts, the sparse data distribution will cause each expert to get less diverse tokens. The expert layer could not get adequately trained. Also, the amount of data is insufficient to match the growing number of parameters, which is also the main reason for overfitting. In order to solve the problems of conventional MoE, our MoEC allows each expert to get more rich and diverse tokens. We impose variance-based constraints on the routing stage, aiming to make neighbor experts have similar routing probabilities for input tokens, thus forming expert clusters prone to tokens with similar hidden states. In MoEC, experts will get a more diverse set of similar input tokens by ``sharing'' input tokens with other experts in the cluster.

Compared with previous work related to MoE, our training objective added an extra term - cluster loss. The overall training objective is to minimize:

\begin{equation}
    \mathscr{L}=\mathscr{L}_{task} +  \mathscr{L}_{balance} +  \mathscr{L}_{cluster}
\end{equation}

$\mathscr{L}_{task}$ is determined by the specific task. In our work, we employ the label smoothed cross-entropy loss for neural machine translation, masked language modeling loss for pre-training language model, and negative log-likelihood loss (NLL loss) or mean-squared loss (MSE loss) for GLUE tasks. In the following, we will introduce $\mathscr{L}_{balance}$ and $\mathscr{L}_{cluster}$.

\textbf{Load Balancing Loss.} 
During training, there exists a load imbalance issue between experts~\cite{shazeer2017outrageously,lepikhin2020gshard}: Most tokens are dispatched to a small number of experts, while many other experts do not get sufficiently trained at all. Besides, imbalanced assignments will result in a high computational bottleneck in the MoE layer and thus limit the computational efficiency. We follow the work in~\cite{fedus2021switch} and add the balance loss to the training objective to encourage a balanced load across experts. Given $N$ experts indexed by $i$=1 to $N$, the balance loss is computed as follows:

\begin{equation}
    \mathscr{L}_{balance}=\alpha N \cdot \sum_{i=1}^{N} f_{i} \cdot p_{i}
\end{equation}

where $f_i$ is the fraction of tokens dispatching to expert $i$. We denote the number of tokens dispatched to the $i$-th expert as $Count_i$. Given a batch $B$ with $T$ tokens, $f_i$ = $Count_i / T$. $p_i$ is the fraction of the routing probability allocated for expert $i$ in the batch $B$. It is calculated by averaging the probability of routing token $x$ to expert $i$ in the batch $B$.

\begin{equation}
    p_{i} = \frac{1}{T} \sum_{x \in B} \alpha_i(x)
\end{equation}

where $\alpha_i(x)$is the gating function depicted in Equation 1, which represents the probability of dispatching token $x$ to expert $i$. The balance loss in Equation 4 encourages uniform routing since it would be minimized under a uniform distribution. To control the impact of balance loss in the training process, a hyper-parameter $\alpha$ is applied as a multiplicative coefficient for the loss. Throughout this work, we use an $\alpha= 10^{-2}$  which was sufficiently large to ensure load balancing while small enough not to overwhelm the primary cross-entropy objective.

\textbf{Clustering Loss.} 

In our work, we find the sparse allocation of data severely hinders the adequate training of MoE layers and exacerbates overfitting. In order to allow experts to get rich and diverse tokens to mitigate the impact of sparse allocation, we design the clustering loss. This loss is designed to constrain certain adjacent experts so that they will share similar routing probabilities to tokens, thus forming a cluster-like distribution. For input tokens originally dispatched to the best-matched experts, clustering loss will give them more opportunities to access other experts in the cluster. As a result, experts will be assigned a more diverse set of similar tokens, thus alleviating the problem of sparse allocation.

In MoE models with $N$ experts, the clustering loss will guide experts to form $m$ clusters ($m$ should be divisible by $N$), and each cluster contains $L=\frac{N}{m}$ experts. We use $E_{i}^{j}$ to represent the j-th expert in the i-th cluster, while $p_{i}^{j}$ represents the routing probability allocated for $E_{i}^{j}$ ($i=0,1,...,m-1; j=0,1,...,L-1$). According to the size and number of clusters, $p_{i}^{0},p_{i}^{1},...,p_{i}^{L-1}$ will compose a one-dimensional matrix $\tilde{P_i} \in \mathbb{R}^{L}$ to represent the routing probabilities of the $L$ experts in the i-th cluster, and we denote the mean value of them as $\overline{p_i}$. We define the clustering loss as follows:

\begin{equation}
\begin{aligned}
    \mathscr{L}_{clustering}
    &=\beta N \cdot \emph{C}_{intra} \cdot \emph{C}_{inter} \\
    &=\beta N \cdot \frac{\sum_{i=0}^{m-1}\delta(\tilde{P_i})}{m} \cdot e^{-\mu\frac{\max{\{\overline{p_i}\}}- max_2\{\overline{p_i}\}}{\max{\{\overline{p_i}\}}}}
\end{aligned}
\end{equation}

As can be seen from Equation 6, clustering loss is mainly composed of two parts: the variance-based intra-cluster constraint $\emph{C}_{intra}$ and the difference-based inter-cluster constraint $\emph{C}_{inter}$.

$\delta(\tilde{P_i})=\frac{(p_{i}^{0}-\overline{p_i})^2+(p_{i}^{1}-\overline{p_i})^2+...+(p_{i}^{L-1}-\overline{p_i})^2]}{L} $ represents the variance of the routing probability in the $i$-th cluster. We compute the mean variance of $m$ clusters as the intra-cluster constraint $\emph{C}_{intra}$, which will be minimized when the routing probabilities of experts within the same cluster are balanced.

Besides, we use $\emph{C}_{inter}$ to measure the probability difference between the dispatched cluster and the sub-optimal cluster. $\max{\{\cdot \}}$ means the max value of $\overline{p_i}$ ($i$=0,1,...,$m$-1) and $max_2\{\cdot \}$ means the second max value. $\emph{C}_{inter}$ will be minimized when the probability of a token being dispatched to a suboptimal cluster is low. $\mu$ is the coefficient used to control the value of $\emph{C}_{inter}$. When we set $\mu=0$, the probability difference between clusters will not be considered. We could also set $\mu$ to a non-zero value to activate $\emph{C}_{inter}$. We will conduct in-depth experiments and analysis on it in the Experiments chapter.

To minimize clustering loss, the probability distribution within the cluster should be uniform, and the probability difference between the clusters should be more apparent (optional). In the initial training steps, the variance among experts will be very high, so the clustering loss will dominate the optimization and guide the rapid formation of expert clusters. When the intra-cluster variance is stable, the clustering loss will become relatively small to maintain the expert clusters. Similar to the practice in balance loss, a hyper-parameter $\beta$ is applied. The value of the $\beta$ should be relatively small, because a large $\beta$ means a strong clustering constraint, thus making experts in the cluster too similar. It will cause these experts to lose their characteristics, and the contributions of multiple similar experts are only approximately equal to one expert. In our work, we set the value of $\beta$ as $10^{-2}$ by default. Experiments on the selection of $\beta$ values could be found in Appendix A. 


\subsection{Cluster-level expert dropout}

When applying large-scale MoE models on tasks with limited data, over-fitting issues naturally arise. Previous MoE-related work~\cite{raffel2019exploring,fedus2021switch} used dropout~\cite{srivastava2014dropout} at each layer to prevent overfitting. Here, cluster-level expert dropout acts as a regularization technique completely different from traditional dropout. It does not drop parameters, but drops some experts in the cluster, which makes the dispatching of tokens more random.

\textbf{Implementation in clusters.}
First, our cluster-level expert dropout works at the routing stage, so it will only be implemented at expert layers. For experts in a cluster, we randomly drop some of them by deleting the expert ids from the candidate expert list when calculating the routing probability. Thus, the corresponding experts will be ignored in the routing stage. Assume the dropout rate as $\gamma$, only the remaining $N(1-\gamma)$ experts will participate in the calculation of routing probability during training. The dimension of the matrix $P$ will decrease from $\mathbb{R}^{N}$ to $\mathbb{R}^{N\cdot(1-\gamma)}$. All clusters implement the dropout simultaneously. It allows tokens to have more opportunities to be dispatched to other experts in the same cluster, instead of being repeatedly dispatched to the expert with the highest probability. From another perspective, each expert will receive more diverse tokens without adding training data.

\begin{figure}
    \centering
    \includegraphics[width=\linewidth]{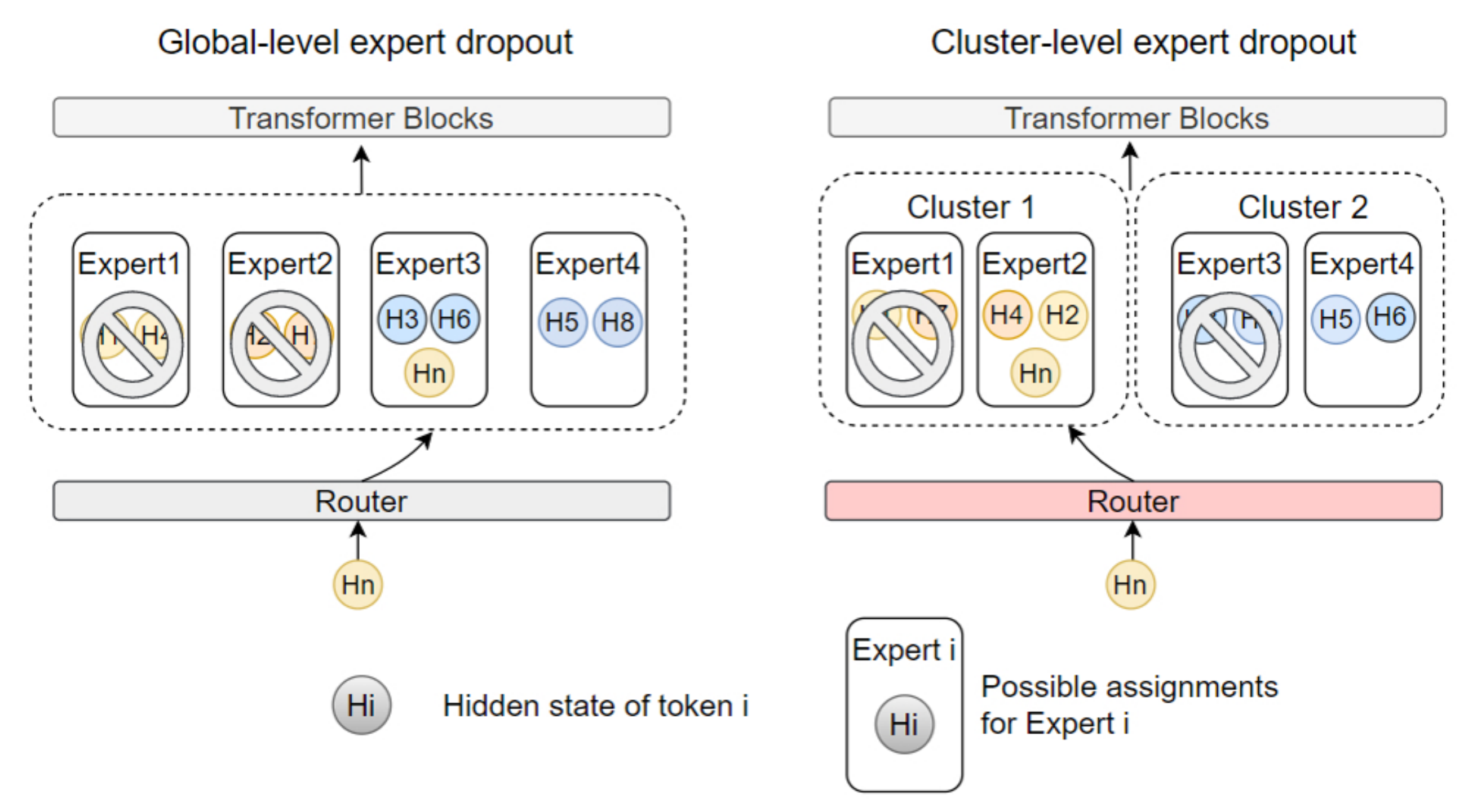}
    \caption{Illustration of global-level expert dropout and cluster-level expert dropout. The similarity between hidden states $H_i$ is represented by the color.} 
    \label{fig:cluster_level}
\end{figure}

\textbf{Cluster-level expert dropout vs Traditional expert dropout.} 

Traditional expert dropout is recommended in~\citet{fedus2021switch}. It is a dropout technique~\cite{srivastava2014dropout} to regularize MoE models, which acts on the feed-forward layer to reduce overfitting caused by too many parameters. By setting a relatively small dropout rate at non-expert layers (0.1), expert dropout increases the dropout rate by an explicit amount at the interim feed-forward computation at each expert layer (0.4). Our expert dropout acts completely different from it. We perform random dropout on the candidate list of experts during the routing stage. It does not reduce the number of parameters during training but allocates tokens more diversely and flexibly. While traditional expert dropout is usually used for fine-tuning on downstream tasks, our cluster-level expert dropout is a general regularization mechanism with strong generality. In addition, our dropout can be applied together with Fedus' expert dropout, and they can work together to improve the performance of MoE.

\textbf{Why cluster-level is better?}

It is natural to think that expert dropout could be implemented at the global level, which provides more opportunities for tokens to access other sub-optimal experts. But for global-level expert dropout, as shown in Figure~\ref{fig:cluster_level}, if a random dropout happens to drop suitable experts, tokens may be dispatched to less relevant experts. Inappropriate dispatching may negatively impact the learning of experts.

In MoEC, We address this problem by exploiting the cluster-like structure and design a cluster-level expert dropout. Cluster-level dropout could give tokens the option to be randomly re-dispatched while confining the routing results to a more reasonable range. No matter how random the dropout is implemented, tokens will always be dispatched to experts with similar routing probability.

\section{Experiments}

\begin{table*}[htbp]
  \centering
  \caption{\textbf{The performance on machine translation and GLUE tasks for baselines and MoEC models.} WMT-14 is measured on the test set, while GLUE tasks are measured on the development sets. We report the average results by a set of seeds (see Appendix C). All experiments are conducted with 64 experts.}
  \resizebox{\linewidth}{!}{
  \begin{tabular}{lcccccccccc}
    \toprule
    & \multicolumn{1}{c}{\textbf{NMT}} & \multicolumn{8}{c}{\textbf{GLUE Tasks}}\\
    \\\cmidrule(lr){2-2}\cmidrule(lr){3-10}
    & \textbf{WMT14 En-De}&\textbf{MNLI}&\textbf{CoLA}&\textbf{SST-2}&\textbf{QQP}&\textbf{QNLI}&\textbf{MRPC}&\textbf{STS-B}&\textbf{GLUE Avg} \\
    \midrule
    Dense &  27.10 &85.97&57.10&92.87&91.20&92.23&87.50 &89.18 &85.16\\  
    MoE Baseline &  30.59 & 87.27& 75.60 & 93.30& 91.37 & 92.33 & 86.30 &88.28 &87.78 \\ 

    \midrule
    MoEC (w/o expert dropout)& 32.21   &87.37 & 75.93& \textbf{93.43} & \textbf{91.45} & 92.40 & 88.07 &89.11 &88.25\\ 
    MoEC & \textbf{32.50}    & \textbf{87.37}&\textbf{76.80} & 93.37& 91.40& \textbf{92.45} & \textbf{88.23} &\textbf{89.24} &\textbf{88.41}\\ 
    \bottomrule
  \end{tabular}
  \label{tab:main_results}}
\end{table*}

We name our model MoEC (Mixture of Expert Clusters), and evaluate the performance on bilingual machine translation and natural language understanding tasks. We use the X-MoE model from~\citet{chi2022representation} as our backbone architecture, which has shown better performance than prior MoE models such as Switch Transformers~\cite{fedus2021switch} on widely-used cross-lingual understanding benchmarks.

\subsection{Evaluation Dataset}

\textbf{WMT 2014 English-to-German} Ninth Workshop on Statistical Machine Translation (WMT 2014) releases a collection of datasets used in shared tasks including machine translation. We add additional news-commentary-v12 data from WMT-17 for training and validation. The total training data contains 3.96M English-to-German sentence pairs.


\noindent\textbf{GLUE} General Language Understanding Evaluation~\cite{wang2018glue} benchmark is a collection of tools for evaluating the performance of models across a diverse set of existing NLU tasks, including MNLI~\cite{williams2017broad}, CoLA~\cite{warstadt2019neural}, SST-2~\cite{socher2013recursive}, QQP, QNLI~\cite{rajpurkar2016squad}, MRPC~\cite{dolan2005automatically} and STS-B~\cite{cer2017semeval}. We do not perform experiments on RTE because previous work~\cite{chen2022task} demonstrated that MoE is not suitable for this task. It is worth mentioning that we will pre-train our model on the BooksCorpus~\cite{zhu2015aligning} and English Wikipedia corpus~\cite{wikidump} for 120k steps before fine-tuning on GLUE tasks.

\subsection{Experiments Setup}

\textbf{Model Architecture}
For our MoEC and all baseline models, we follow the recommended settings in~\cite{vaswani2017attention} and use Transformer-big as the unified backbone architecture on WMT 2014 English-German translation task. For GLUE tasks, we use Transformer-base as the backbone architecture.

For MoE layers, we apply the 64-expert MoE model with 3 FFN sub-layers in the 3rd encoder block and 3rd decoder block. A more detailed model hyper-parameters could be found in Appendix B.

\noindent\textbf{Baselines}

We conduct two baselines in our experiments. The first is \textbf{dense transformer}~\cite{vaswani2017attention}. For another, we follow the work in~\cite{chi2022representation} and apply X-MoE as our~\textbf{MoE baseline}. It could serve as a strong baseline that shows better performance than Switch Transformer~\cite{fedus2021switch} on widely-used cross-lingual understanding benchmarks. The MoE baseline estimates routing scores between tokens and experts on a low-dimensional hypersphere and adds a learnable temperature scalar in the gating function. For a fair comparison, the two baseline methods are built with the same setting as MoEC, which could be found in Appendix B.

\noindent\textbf{MoEC Hyper-parameters}
For MoEC, several unique hyper-parameters are introduced. For clustering loss, we set $\beta$ to $10^{-2}$ according to the experiment results (see Appendix A) and set $\mu=0$ by default. For cluster size (the number of experts in a cluster) and expert dropout rate, we will have detailed related experiments in the following sections.

\noindent\textbf{Training Hyper-parameters}
For a fair comparison, the dense model, MoE baseline model, and MoEC model share the same training hyper-parameters. All models are trained with the Adam optimizer~\cite{kingma2014adam} ($\beta_1=0.9,\beta_2=0.98$). The learning rate is set $5e^{-4}$ with 4000 warm-up steps and inverse square root scheduler~\cite{raffel2019exploring}. Batch size, training steps, and dropout rate are set by different tasks, which are recorded in Appendix C.


\subsection{Experiments results}

We train dense models, baseline MoE and MoEC models on several widely-used evaluation tasks, and the results are shown in Table~\ref{tab:main_results}. Compared with dense models, MoE models exhibit significant performance improvements, which benefit from the large model capacity. Besides, MoEC could bring notable improvement over the MoE baseline without applying the dropout strategy to experts. On WMT-14, it gives a 1.62 BLUE score boost. The advantage could be attributed to the clustered distribution of experts, which endows experts with more diverse and appropriate training samples. Moreover, with the application of the cluster-level expert dropout strategy, the performance of MoEC will be further improved.

\begin{figure}
    \centering
    \includegraphics[width=\linewidth]{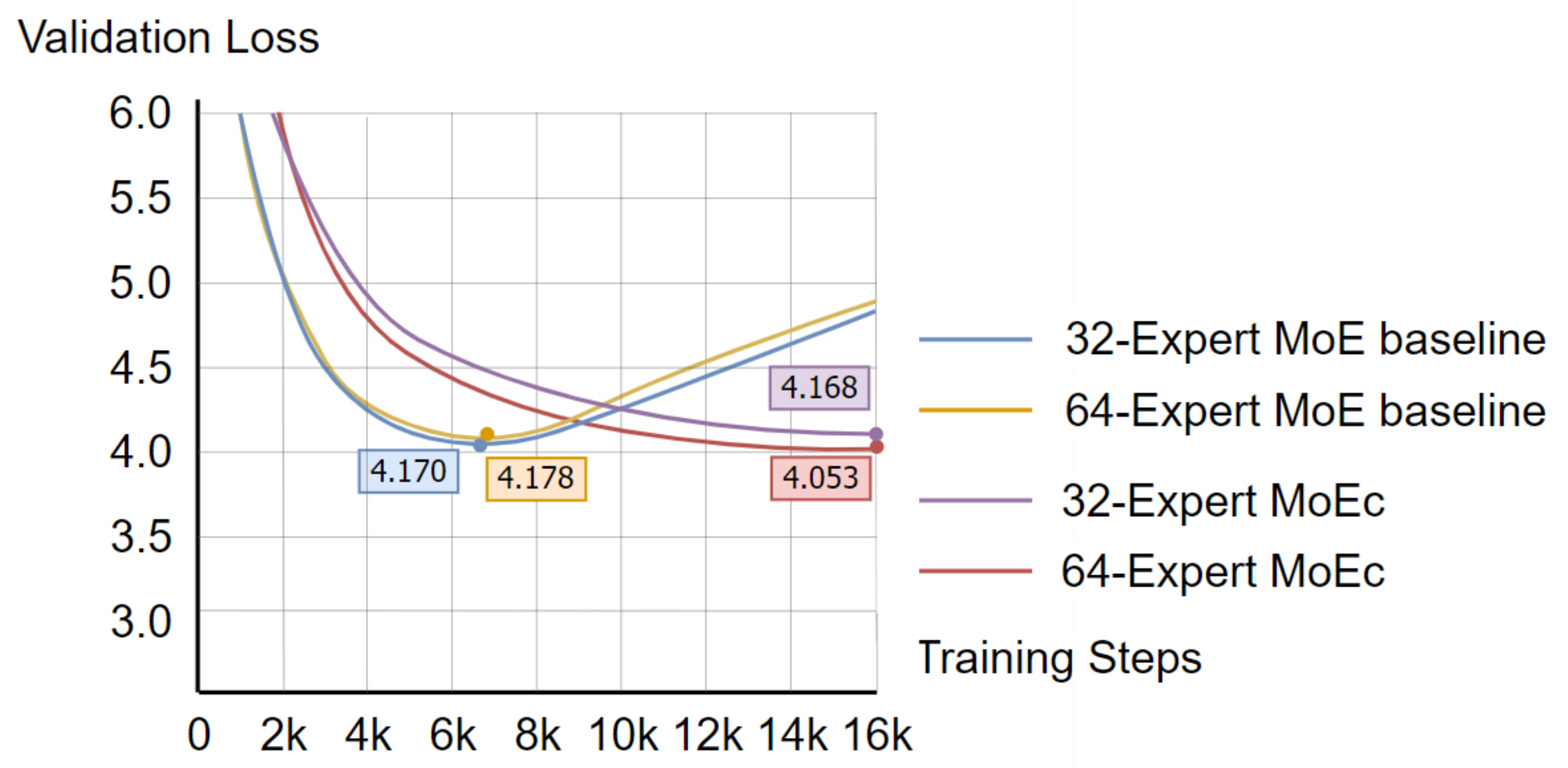}
    \caption{Loss curves on the WMT-14 validation set. All experiments are conducted with 64 experts for a fair comparison. The validation loss that rises with increasing training steps indicates the overfitting phenomenon. Our MoEC shows excellent ability to mitigate overfitting.} 
    \label{fig:val_loss}
\end{figure}

As shown in Figure~\ref{fig:val_loss}, the MoE baseline severely suffers from overfitting on WMT-14, while our MoEC shows excellent ability to mitigate overfitting. The overfitting phenomenon on the validation set is almost eliminated, and the validation loss is relatively lower. It shows that when our MoEC solves the sparse allocation of data, each expert could get more abundant and diverse training samples. In this way, the training data of each expert is kept sufficient, thereby alleviating the phenomenon of overfitting. Furthermore, we found that MoEC converges slightly slower. It is due to the fact that each expert needs to learn from more diverse training samples, which takes more steps to allow the expert to get sufficiently trained.

\subsection{Detailed analysis of expert clusters}

\begin{figure}
    \centering
    \includegraphics[width=\linewidth]{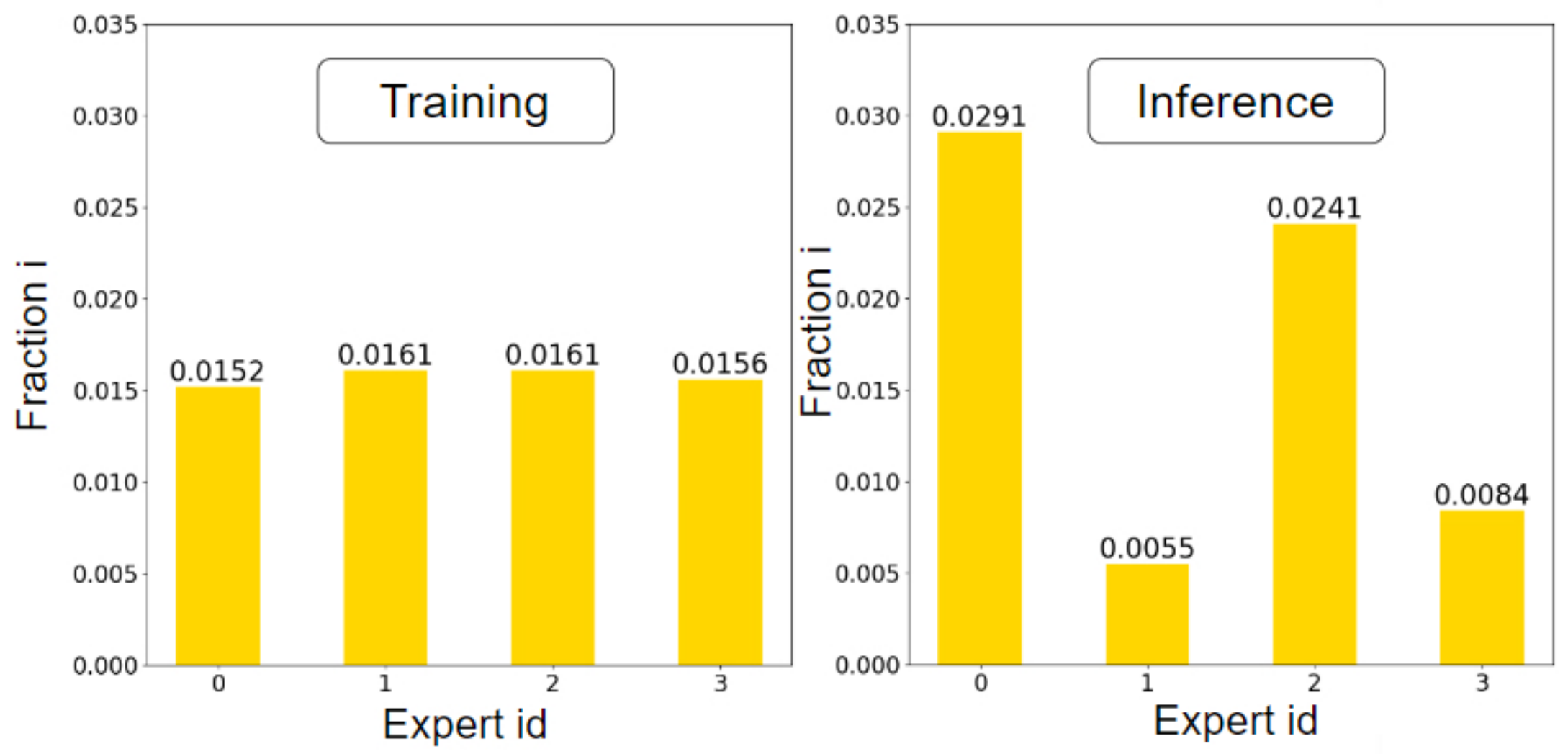}
    \caption{Fraction of tokens dispatched to Expert 0$\sim$3 (i.e. $f_i$ mentioned above) of 64-expert MoEC (cluster size = 4) during training and inference. The graph on the left represents the fraction of tokens dispatched to cluster 0 during training, while the right shows the fraction of tokens dispatched to cluster 0 during inference.} 
    \label{fig:MoEC}
\end{figure}

Next, we conduct a detailed analysis of expert clusters. Figure~\ref{fig:MoEC} shows the fraction of tokens dispatched to cluster 0 (expert 0$\sim$3) during training and inference. During training, the experts in cluster 0 get similar input tokens, which are affected by balance loss and clustering loss. During inference, the routing probabilities of experts in the cluster vary, which indicates that they still retain their own characteristics. They learn more fine-grained knowledge, which is the advantage of multiple similar experts compared to a single expert. For WMT14, the BLUE score of MoE with 16 experts is 30.49, while the BLUE score of MoE with 16 clusters (cluster size=4) is 32.16. It shows that multiple similar experts have an obvious advantage over a single expert.

The cluster size also has a critical impact on the learning of MoEC, so we conduct experiments on different cluster sizes. As depicted in Table~\ref{tab:cluster-size}, the best performance is obtained when cluster size = 8. Compared to the MoE baseline with 64 experts, expert clusters could bring about a 1.62 BLUE scores improvement.
When the cluster size is relatively small, the data shared among experts will be less, and the improvement brought by MoEC will not be fully exploited. As a special case, when cluster size=1, a single expert could not be called a cluster, and MoEC is equivalent to MoE baseline. When the cluster size is large, the data shared among experts will increase, but the similarity and correlation of these data will become lower, which will lead to an adverse impact on the ``professionalism" of each expert. When we expand the cluster size to 16, the performance of MoEC is even lower than that of the MoE baseline, which means that an excessively large cluster size will suppress the advantages of MoE structure and hurt the performance.

\begin{table}[htbp]
  \centering
  \caption{\textbf{The performance of MoEC with different cluster sizes on WMT-14.} All experiments were conducted with 64 experts. For a fair comparison, all methods do not employ the dropout on experts.}
   \begin{tabular}{ccc}
    \toprule
    \textbf{Cluster size}& \textbf{Number of clusters} & \textbf{BLEU}\\
    \midrule
    1 &64& 30.59 \\
    4 &16& 32.16 \\
    8 &8 & \textbf{32.21} \\
    16&4 & 29.98 \\

    \bottomrule
    \end{tabular}
  \label{tab:cluster-size}
\end{table}

\subsection{Expert dropout: Cluster-level vs global-level}

\begin{table}[htbp]
  \centering
  \caption{\textbf{Cluster-level vs global-level expert dropout on WMT-14.} All experiments are conducted on the 64-expert MoEC and cluster size = 8. Under this setting, the BLUE score of MoEC without expert dropout is 32.21.}

   \begin{tabular}{ccc}
    \toprule
    
    \textbf{Dropout rate}&\textbf{cluster-level}&\textbf{global-level}  \\
    \midrule
    0    & 32.21 & 32.21 \\
    0.25 & 32.32 & 31.88 \\  
    0.5  & \textbf{32.50} & 31.53 \\ 
    0.75 & 32.02 & 29.73 \\ 

    \bottomrule
    \end{tabular}
  \label{tab:dropout}
\end{table}

In Table~\ref{tab:dropout}, we experiment on WMT-14 with the cluster-level expert dropout rate. We find that cluster-level dropout could enhance the generalization performance of MoEC. Such a regularization method could bring a 0.29 BLUE scores improvement for MoEC. Experimental results show that 0.5 is a good choice for the dropout rate.
Besides, it is obvious that global-level expert dropout will hurt the performance.


For cluster-level expert dropout, when dropping the best-matched expert for input tokens, the routing decision will still be made among the rest experts in the cluster. Regardless of how the dropped experts are selected, there will always be experts left in each cluster. It ensures that suitable experts are always available. But for the global-level one, due to the random distribution of experts, if all matched experts are dropped, the token will be routed to an inappropriate expert. It could cause experts to be distracted by low-relevant data, thus negatively impacting the learning of knowledge. Take Figure~\ref{fig:cluster_level} as a simple example (with setting the dropout rate to 0.5). For global-level expert dropout, when both expert1 and expert2 are dropped, then $H_n$ will only be dispatched to expert3 or expert4. This inappropriate allocation could hurt the performance of the model.


\subsection{Role of the inter-cluster constraint coefficient $\emph{C}_{inter}$}

\begin{figure}
    \centering
    \includegraphics[width=\linewidth]{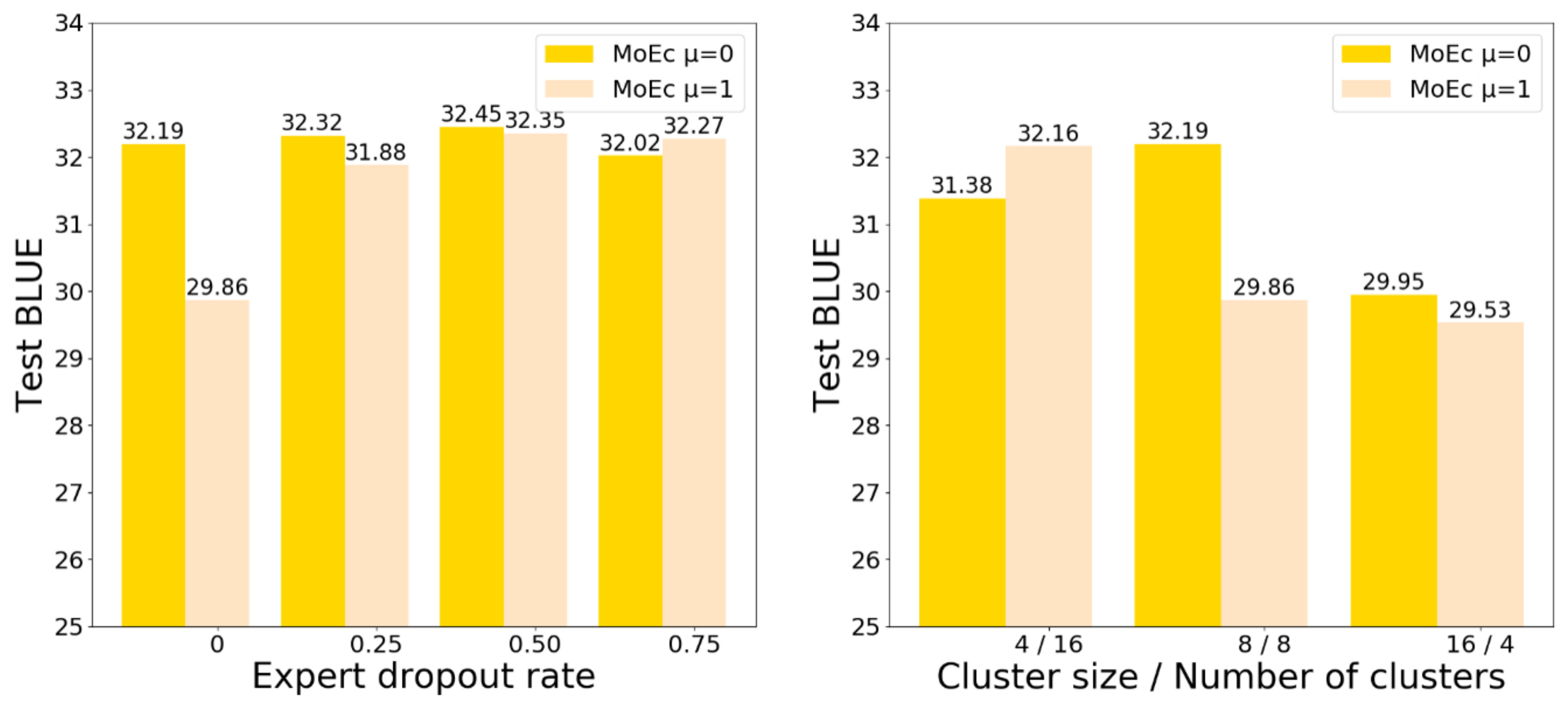}
    \caption{Two sets of experiments on the inter-cluster constraint coefficient $\emph{C}_{inter}$. All experiments are performed on WMT14 En-De. The figure on the left is about experiments with different expert dropout rates (cluster size=8), and The figure on the right is about experiments with different cluster sizes (without expert dropout). }
    \label{fig:vs}
\end{figure}

We further explore whether the inter-cluster constraint coefficient $\emph{C}_{inter}$ (in Equation 6) will help improve performance. As depicted in Figure~\ref{fig:vs},  when dropout=0.75 or cluster size=4, setting $\mu$ to 1 will get better results. In other cases, it is better not to apply inter-cluster constraints by setting $\mu$ to 0.

When there are sufficient experts in the cluster, it is better not to use the inter-cluster constraint by setting $\mu$ to 0. Intra-cluster constraints have already made other experts in the cluster have higher routing probabilities, while inter-cluster constraints will further widen the routing probability gap between clusters. This will cause the entropy of the routing probability distribution to be too small, which is not conducive to the learning of the gated network.

We find that the inter-cluster constraint will benefit MoEC when the cluster size is small or the expert dropout rate is high. In this case, the number of experts in the cluster is small, and the intra-cluster constraint alone is not enough to form a globally reasonable routing probability distribution, so the assistance of constraints between clusters is needed.

\subsection{Raising the upper bound of MoE}


\begin{table}[htbp]
  \centering
  \caption{\textbf{Results of scaling up MoEC.}}
   \begin{tabular}{cccc}
    \toprule
    \textbf{Expert num} &\textbf{MoE baseline}&\textbf{MoEC}&\textbf{Benefits}\\
    \midrule
    16  & 30.49 &30.50&+0.01\\
    32  & \textbf{30.81} &30.84&+0.03\\
    64  & 30.59 & \textbf{32.50}&+1.91\\
    128 & 30.21 & 32.40&\textbf{+2.19}\\
    \bottomrule
    \end{tabular}
  \label{tab:scale-up}
\end{table}

In general, a higher number of experts means higher model capacity and better performance. However, for tasks with limited data, there exists a performance upper bound on scaling up MoE models. We take a deep dive into the ability of MoEC to raise the upper bound. As shown in Table~\ref{tab:scale-up}, for the MoE baseline, expert = 32 is the upper bound, which means that continuing to increase the number of experts will not bring any gain to the model. Our MoEC not only has a performance advantage over the MoE baseline with the same number of experts, but also improves the upper bound from 32 to 64. 

With the increase of experts, our MoEC could bring more gains. It is because MoEC could fully show its promising ability to solve severe overfitting and sparse allocation problems. With the mitigation of the above two problems, the superiority of the large-scale MoE model will be better exerted, thereby achieving the improvement of the upper bound of MoE models. With the help of MoEC, we could try to build sparse models with more experts.

\section{Conclusion}

In our work, we point out the overfitting and the sparse data allocation problems of large-scale MoE models and propose a novel training strategy - MoEC to convert experts into clusters. Each expert could get more abundant and diverse training samples. In this way, the training data of each expert is kept sufficient, thereby alleviating overfitting. We also propose the cluster-level expert dropout to realize regularization. We conduct experiments on machine translation and natural language understanding tasks. Experiment results show MoEC could improve performance and alleviate problems caused by scaling up experts without changing the model structure and routing strategy. The superiority of the large-scale MoE model will be better exerted by MoEC, thereby raising the upper bound of MoE models. With the help of MoEC, we could try to build sparse models with more experts.

\bigskip



\bigskip

\bibliography{aaai22}

\appendix
\section{Appendix}
\subsection{A Selection of the value of $\beta$}

\begin{table}[htbp]
  \centering
  \caption{\textbf{The performance of MoEC with different $\beta$ coefficients on WMT-14.} All experiments are conducted with 64 experts. The cluster sizes=8, and expert dropout rate=0.25.}
   \begin{tabular}{cc}
    \toprule
    \textbf{Value of $\beta$} &  \textbf{MoEC}  \\
    \midrule
    1e-3 & 32.21\\
    5e-3 & 32.17\\
    1e-2 & \textbf{32.32}\\
    5e-2 & 31.21\\
    \bottomrule
    \end{tabular}
  \label{tab:beta coef}
\end{table}

Table~\ref{tab:beta coef} presents the experiments on selecting the best value of $\beta$. MoEC works best when $\beta$ is set to 1e-2. And when the beta value is too large, the performance of MoEC drops significantly, which confirms our analysis in the main text. Based on the results, we uniformly set the value of $\beta$ as $10^{-2}$ as a default in all experiments above.

\subsection{B Architecture parameters}

Table \ref{tab:arch_paras} presents the architecture parameters for different tasks.

\begin{table}[htbp]
  \centering
  \caption{\textbf{Architecture parameters for all tasks}}
  \resizebox{\linewidth}{!}{
   \begin{tabular}{ccc}
    \toprule
     - &\textbf{WMT-14 En-De}  &\textbf{Pre-train\&GLUE} \\
    \midrule
    Transformer blocks & 12  &12\\
    Attention heads & 16  &12\\
    Encoder/Decoder embedding & 1024  &768\\
    FFN embedding & 4096  &3072\\
    \midrule
    Experts & [16,32,64,128]&[16,32,64,128]\\
    Routing dimension & [8,16,32,64]  & [8,16,32,64] \\
    MoE layers & 2&1\\
    Sub-layers & 3&3\\

    \bottomrule
    \end{tabular}}
  \label{tab:arch_paras}
\end{table}

\subsection{C Training hyper-parameters}

Table~\ref{tab:train_paras} presents the training hyper-parameters for WMT-14 and pre-training. Table~\ref{tab:blue_paras} presents the training hyper-parameters on downstream GLUE tasks.

\begin{table}[htbp]
  \centering
  \caption{\textbf{Training hyper-parameters for all tasks}}
  \resizebox{\linewidth}{!}{
   \begin{tabular}{ccccc}
    \toprule
     - &\textbf{WMT-14 En-De}  &\textbf{Pre-train} \\
     \midrule
     Optimizer& Adam&  Adam\\
     Adam $\epsilon$& 1e-6&  1e-6  \\
     Adam $\beta$ & (0.9,0.98)&  (0.9,0.98)\\
     Training Steps& 32k &125k\\
     Batch size& 8k&2k\\
     Maximum learning rate&5e-4&5e-4 \\
     Learning Rate Scheduler& inverse sqrt& inverse sqrt \\
     Warmup steps& 4k  & 4k \\
     Weight decay& 0  &0.01 \\
     Dropout&0.3&0.1 \\
     Attention dropout&0.1&0 \\
     Gradient Clip Norm& 0.1& 0.1 \\
     Label smoothing & 0.1 & - \\
     \midrule
     Capacity factor& 2& 2\\
     MoE dropout&0.4&0 \\
     MoE activation dropout&0.1&0 \\
     balancing coefficient $\alpha$&0.01 &0.01\\
    \bottomrule
    \end{tabular}}
  \label{tab:train_paras}
\end{table}

\begin{table}[htbp]
  \centering
  \caption{\textbf{Training hyper-parameters for GLUE.}}
  \resizebox{\linewidth}{!}{
   \begin{tabular}{ccccccccc}
    \toprule
     \textbf{Hyper-parameters} &\textbf{MNLI} &  \textbf{SST-2} &\textbf{QQP}&\textbf{QNLI} &\textbf{CoLA}&\textbf{STS-B}&\textbf{MRPC}&\textbf{RTE}\\
     
     \midrule
    Batch Size & 32& 32& 32& 32& 32& 32& 32& 32 \\
    Epochs & [3,5]& [3,5]& [3,5]& [3,5]& [3,5,10]& [10,15,20]& [5,10,15,20]& [3,5,10] \\
    Learning rate & [1,2,4]e-5& [1,2,4]e-5& [1,2,4]e-5& [1,2,4]e-5& [1,2,4]e-5& [1,2,4]e-5& [1,2,4]e-5& [1,2,4]e-5\\
    Warm up & 16& 16& 16& 16& 16& 16& 16& 16\\
    Seed & [1,2,3]& [1,2,3]& [1,2,3]& [1,2,3]& [1,2,3]& [2,42,123]& [2,42,123]& [1,2,3]\\
    
    \bottomrule
    \end{tabular}}
  \label{tab:blue_paras}
\end{table}

\end{document}